\documentclass[a4paper,11pt,twocolumn,twoside]{article}
\usepackage[dvips]{graphicx}
\usepackage{tipa}
\usepackage{sepln_en}
\usepackage{fullname}
\usepackage[english]{babel}  
\usepackage{xurl}  
\usepackage[ruled,vlined,linesnumbered]{algorithm2e}
\usepackage[noend]{algpseudocode}
\usepackage{amsmath}

\usepackage{enumerate}
\usepackage[shortlabels]{enumitem}

\usepackage{textcomp}

\DeclareFixedFont{\ttb}{T1}{txtt}{bx}{n}{9} 
\DeclareFixedFont{\ttm}{T1}{txtt}{m}{n}{9}  

\usepackage{color}
\definecolor{deepblue}{rgb}{0,0,0.5}
\definecolor{deepred}{rgb}{0.6,0,0}
\definecolor{deepgreen}{rgb}{0,0.5,0}

\usepackage{listings}

\usepackage[utf8]{inputenc}
\input epsf

\title{BERTIN: \\Efficient Pre-Training of a Spanish Language Model using Perplexity Sampling}

\author{Anonymous}
\author {\textbf{Javier de la Rosa$^{1,2}$}, \textbf{Eduardo G. Ponferrada$^1$}, \textbf{Paulo Villegas$^{1,3}$}, \\
\textbf{Pablo González de Prado Salas$^{1,4}$}, \textbf{Manu Romero$^{1,5}$}, \textbf{María Grandury$^{1}$} \\
$^1$BERTIN Project\\
$^2$National Library of Norway, Mo i Rana, Norway\\
$^3$Telefónica I+D, Madrid, Spain\\
$^4$Foqum, Madrid, Spain\\
$^5$Narrativa, Madrid, Spain\\
\\
\texttt{versae@nb.no}, \texttt{edugp91@gmail.com}, \texttt{paulo.vllgs@gmail.com} \\
\texttt{pablogps86@gmail.com}, \texttt{mrm8488@gmail.com}, \texttt{mariagrandury@gmail.com}}

\seplntranstitle{BERTIN: \\Preentrenamiento eficiente de un modelo de lenguaje en español usando muestreo de perplejidad}

\seplnclave{Modelos de lenguaje preentrenados. Métodos de muestreo. IA dato-céntrica.}

\seplnresumen{El preentrenamiento de grandes modelos de lenguaje generalmente requiere cantidades masivas de recursos, tanto en términos de computación como de datos. Las fuentes web comúnmente usadas, como Common Crawl, pueden contener el suficiente ruido para que el preentrenamiento no sea óptimo. En este trabajo experimentamos con diferentes métodos de muestreo de la versión en español de mC4 y presentamos una técnica novedosa centrada en datos que llamamos \textit{muestreo de perplejidad} y que permite el preentrenamiento de modelos de lenguaje en aproximadamente la mitad de pasos, y con una quinta parte de los datos normalmente necesarios. Los modelos obtenidos logran resultados comparables e incluso superan el estado del arte para ciertas tareas. Nuestro trabajo es una muestra de la versatilidad de los modelos Transformers en cuanto a aprendizaje práctico y allana el camino para que otros equipos pequeños entrenen sus modelos con un presupuesto limitado.}

\seplnkey{Pre-trained Language Models. Sampling Methods. Data-centric AI.}

\seplnabstract{The pre-training of large language models usually requires massive amounts of resources, both in terms of computation and data. Frequently used web sources such as Common Crawl might contain enough noise to make this pre-training sub-optimal. In this work, we experiment with different sampling methods from the Spanish version of mC4, and present a novel data-centric technique which we name \textit{perplexity sampling} that enables the pre-training of language models in roughly half the amount of steps and using one fifth of the data. The resulting models are comparable to the current state-of-the-art, and even achieve better results for certain tasks. Our work is proof of the versatility of Transformers, and paves the way for small teams to train their models on a limited budget.}

\firstpageno{1}

\begin{document}

\setlength\titlebox{22cm} 

\label{firstpage} \maketitle

%

\section{Introduction}

Since the introduction of the Transformer architecture \cite{vaswani17}, the number of parameters in language models and the amount of data used for training them have grown almost linearly over the years \cite{HAN2021}. While estimates suggest that roughly 5GB of English text was used for the original GPT model \cite{radford_improving_2018} and almost 16GB for BERT \cite{devlin18}, subsequent versions like GPT-2, RoBERTa, T5, or GPT-3 scaled the training corpora from 40GB to almost 570GB \cite{Radford2019LanguageMA,liu19,2020t5,gpt3,mesh-transformer-jax}. And this trend seems to be nowhere near an end \cite{fedus2021switch,lieberetal21}.

Most language models are first released for English, for which very large and high-quality training sets exist \cite{pile}. Resources of comparable quality are not always available for other languages, but some do have sufficiently large corpora to train monolingual versions \cite{YUAN202165,xue-etal-2021-mt5}. Regardless, relevant contributions like BERT, XLNet or GPT2 often take years to be available in these languages and, when they do, it is often via multilingual versions which are not as performant as their monolingual alternatives. In this context, a few questions remain unclear regarding the pre-training datasets for high-resource languages. In particular:

\begin{enumerate}[start=1,label={(\bfseries RQ\arabic*):}]
\item How much data is enough to train a well-performing monolingual language model?
\item When more than enough data exist, how to select the documents that enable a more efficient training?
\item How does data quality affect training times?
\end{enumerate}

In order to answer these questions, we explore a technique to sample documents at training time from a large dataset of web crawled content. As the second most-spoken language in the world by native speakers\footnote{Over 470 million speakers. ``What are the top 200 most spoken languages?". Ethnologue. \url{https://www.ethnologue.com/guides/ethnologue200}. Retrieved 2022-02-20.}, we chose Spanish as our testing language, and RoBERTa as our language model architecture. In this work, we consider the hypothesis that sampling methods might help reduce training-data size and training times, without a noticeable impact on the performance of the final model.

\section{Data and Methods}
At the time of performing our experiments, no RoBERTa models were publicly available for Spanish. Models in monolingual Spanish are generally hard to come by and, when they do, they are often trained on proprietary datasets and with massive resources \cite{padro2012freeling,gutierrezfandino2021spanish}. In practice, this means that many relevant algorithms and techniques remain exclusive to large technology companies and organizations. 

\subsection{Spanish \texttt{mC4}}

The \texttt{mC4} dataset is a multilingual variant of \texttt{C4}, the `Colossal, Cleaned version of Common Crawl's web crawl corpus'. While \texttt{C4} was used to train the T5 text-to-text Transformer models, \texttt{mC4} comprises natural text in 101 languages drawn from the public Common Crawl web-scrape and was used to train mT5, the multilingual version of T5 \cite{xue-etal-2021-mt5}.

The Spanish portion of \texttt{mC4} (\texttt{mC4-es}) contains about 416 million documents and 235 billion words in approximately 1TB of uncompressed data\footnote{416,057,992 documents and 235,303,687,795 words}.

\subsection{Perplexity sampling}\label{perplexity_sampling}

The large amount of text in \texttt{mC4-es} makes training a language model in constrained environments very challenging. To overcome this limitation, we explored sampling methods to create subsets of \texttt{mC4-es} that would enable the training of language models with roughly one fifth of the data (around 200GB of data containing 50M documents) at approximately half the training steps used to pre-train a regular RoBERTa-base.

In order to adequately build this subsets of data, we decided to leverage a technique we call \textit{perplexity sampling}, and whose origin can be traced to the construction of CCNet and their high-quality monolingual datasets from web-crawled data \cite{wenzek2019ccnet,conneau2019unsupervised}. In their work, they suggest the possibility of applying fast language models trained on high-quality data such as Wikipedia to filter out texts that deviate too much from correct expressions of a language. For each of the 100 languages with the largest Wikipedia, the authors also trained and released a Kneser-Ney model \cite{ney1994structuring} as implemented in the KenLM library \cite{heafield2011kenlm}. However, they decided not to remove content based on the KenLM score because they considered that some of it could be useful for specific downstream applications. Moreover, they picked perplexity thresholds for each language and split the corpus in 3 parts of equal size. They did notice that the part with higher perplexity values achieved slightly better results. This is fundamentally different from our approach. On one hand, we do not perform filtering but sampling, which are two distinct operations with different purposes, contexts, and goals. On the second hand, we do not split the corpus in equally sized parts, but incorporate the notion of statistical quartiles to bias against poor quality documents.

In order to test our hypothesis, we first calculated the perplexity of each document in a random subset (roughly a tenth of the data) of \texttt{mC4-es} and extracted their distribution and quartiles (see Figure~\ref{figure2}). 

\begin{figure}[h]
  \centering
  \includegraphics[width=7cm,clip]{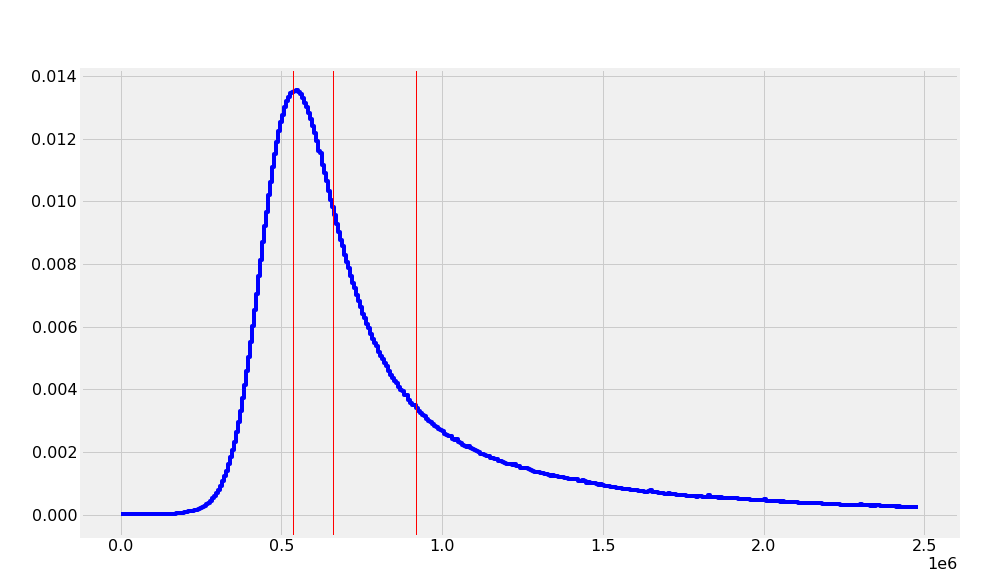}
  \caption{Perplexity distribution (blue) and quartiles (red) of 44M documents of \texttt{mC4-es}.}
  \label{figure2}
\end{figure}

The probability $p(w_{n} \mid w_{1}^{n-1})$ (\ref{eq0}) of a word in backoff-smoothed models such as Kneser-Ney where $w_1^n$ is a context $n$-gram, is based on the observed entry with longest matching history $w_f^n$, with backoff penalties given as $b(w_{i}^{n-1})$ by an already-estimated model.

\begin{equation}
\label{eq0}
\resizebox{.26\hsize}{!}{$p\left(w_{n} \mid w_{1}^{n-1}\right)$} = \resizebox{.26\hsize}{!}{$p\left(w_{n} \mid w_{f}^{n-1}\right)$} \prod_{i=1}^{f-1} \resizebox{.18\hsize}{!}{$b\left(w_{i}^{n-1}\right)$}
\end{equation}

KenLM models are part of the Kneser-Ney family of models. In KenLM, the perplexity score for a given sentence is based on the probabilities of its constituent words as computed by the model (\ref{eq00}).

\begin{equation}
\label{eq00}
pp(s) = p(w_1, w_2, ..., w_N)^{-1/N}
\end{equation}

Since we were aiming at speed, we decided to skip the SentencePiece tokenization step in the calculation of the perplexity. In contrast to \namecite{wenzek2019ccnet} and \namecite{conneau2019unsupervised}, we feed the raw, unnormalized strings, line by line, to a 5-gram KenLM model trained on the Spanish Wikipedia. Thus, the perplexity is calculated as in (\ref{eq1}), where $W$ is a document with $L$ lines, and $KenLM(W_i)$ returns the score for the $i$-th line in the document.

\begin{equation}
\label{eq1}
    pp(W) = 10^{\frac{-\sum_{i=1}^{L}{KenLM(W_i)}}{L}}
\end{equation}

%



With the extracted perplexity values, we created two functions to oversample the central quarters of the perplexity distribution with the goal of biasing against documents whose perplexity is either too small (short, repetitive texts) or too long (potentially poor quality), and then we compared them to a random sampling. The first function is a step function (\texttt{Stepwise}) that oversamples the central quarters while subsampling the rest (\ref{eq2}). For perplexity values in the two central quarters of the distribution, it gives larger frequencies that are inversely proportional to their respective quartile ranges. For values of perplexity outside the central quarters, it gives lower frequencies inversely to the quartiles. As a result, the step function generates a piecewise transformation of the perplexity distribution. We adjusted $\alpha$ to be roughly a 10\% of $Q_3$ to balance out the high perplexity values that result from skipping the SentencePiece tokenization\footnote{We did not assess the impact of using SentencePiece during the original experiments. However, we generated post-hoc the distributions for a few thousand documents with and without this tokenization method. When using SentencePiece, the raw values of perplexity were significantly lower, and the spread was a bit higher than without it. Nonetheless, the distributions were very similar in shape.}.

\begin{align}
\label{eq2}
\resizebox{.2\hsize}{!}{$p_{stepwise}(W)$} = \left\{ \begin{array}{cr} 
                \frac{\alpha}{Q_1} & \resizebox{.2\hsize}{!}{$pp(W) \leq Q_1$} \\
                \frac{\alpha}{Q_2 - Q_1} & \resizebox{.3\hsize}{!}{$Q_1<pp(W) \leq Q_2$} \\
                \frac{\alpha}{Q_3 - Q_2} & \resizebox{.3\hsize}{!}{$Q_2<pp(W) \leq Q_3$} \\
                \frac{\alpha}{Q_3} & \resizebox{.2\hsize}{!}{$pp(W) > Q_3$} \\
                \end{array} \right.
\end{align}

The second approach weights the perplexity distribution using a Gaussian-like function, where $\tilde{X}$ represents the median of the perplexity distribution ($Q_2$), to smooth out the sharp boundaries of the \texttt{Stepwise} function and to give a better approximation to the desired underlying distribution. Thus, the probability of keeping a given document $W$ is given by (\ref{eq3}).

\begin{equation}
\label{eq3}
    p_{gaussian}(W) = \alpha \cdot e^{-\frac{1}{\beta}\left(\frac{pp(W) - \tilde{X}}{\tilde{X}}\right)^2}
\end{equation}

We adjusted the $\alpha$ parameter of the \texttt{Stepwise} function, and the $\alpha$ and $\beta$ (spread) parameters of the Gaussian function to be able to extract roughly 50M documents from the 416M in \texttt{mC4-es} (see Figures \ref{figure3} and \ref{figure4}). As a baseline, we also sampled randomly \texttt{mC4-es} up to 50M documents. In terms of sizes, we went down from 1TB of raw data to 200GB. However, when these parameters were applied to the validation split they resulted in too few examples (fewer than 400k documents). Therefore, for validation purposes, we extracted 50k documents at each evaluation step from our own training dataset. Crucially, those documents were then excluded from training, so as not to validate on previously seen data. Figure \ref{figure5} shows the actual perplexity distributions of the generated 50M subsets for each of the executed sampling procedures. Random sampling exhibited the same perplexity distribution of the underlying true distribution, as can be seen in Figure \ref{figure6}.

\begin{figure}[h]
  \centering
  \includegraphics[width=7cm,clip]{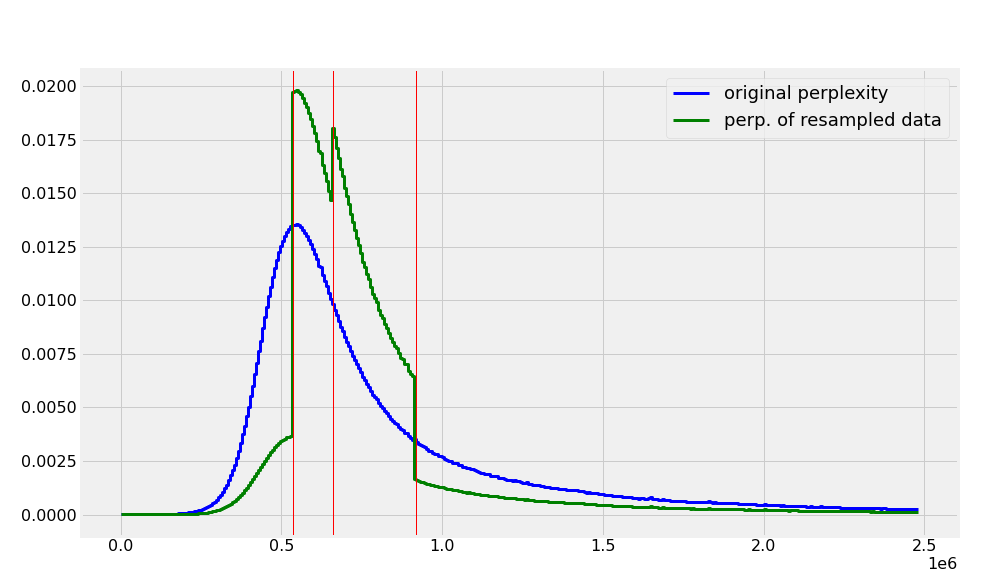}
  \caption{Expected perplexity distributions of the sample \texttt{mC4-es} after applying the \texttt{Stepwise} function.}
  \label{figure2}
\end{figure}

\begin{figure}[h]
  \centering
  \includegraphics[width=7cm,clip]{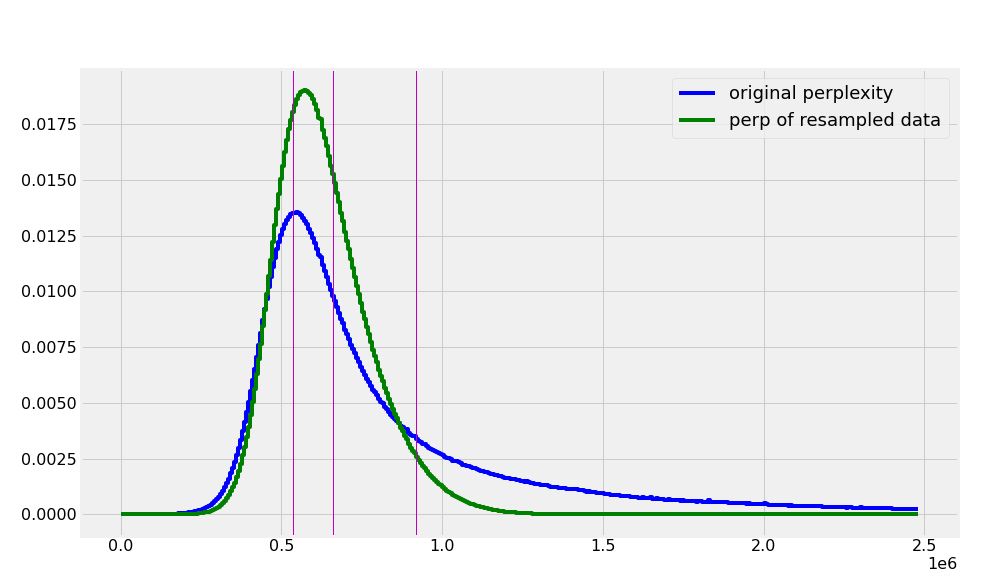}
  \caption{Expected perplexity distributions of the sample \texttt{mC4-es} after applying the \texttt{Gaussian} function.}
  \label{figure3}
\end{figure}

\begin{figure}[h]
  \centering
  \includegraphics[width=7cm,clip]{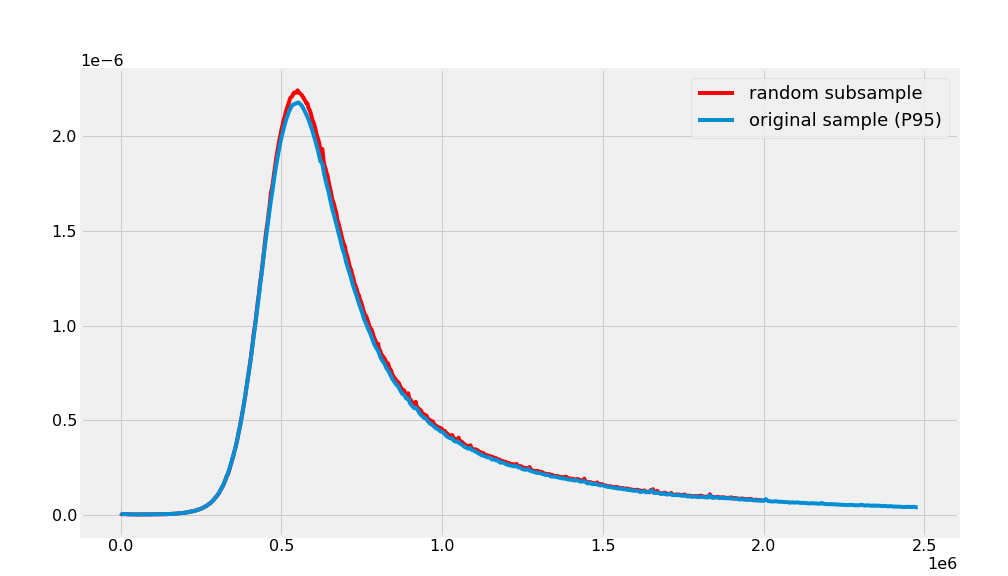}
  \caption{Experimental perplexity distribution of the sampled \texttt{mC4-es} after applying Random sampling.}
  \label{figure5}
\end{figure}

\begin{figure}[h]
  \centering
  \includegraphics[width=7cm,clip]{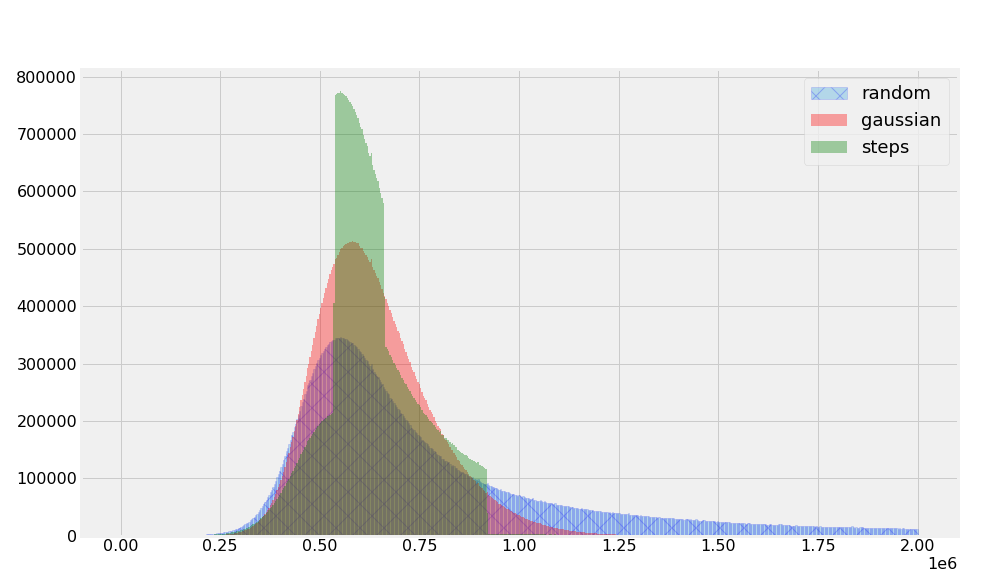}
  \caption{Experimental perplexity distributions of the sampled \texttt{mC4-es} after applying \texttt{Gaussian} and \texttt{Stepwise} functions, and the random control sample.}
  \label{figure4}
\end{figure}

A quick t-SNE plot (see Figure \ref{figure6}) seems to suggest that the distribution is uniform for the different topics and clusters of documents. The plot was generated using a distilled version of multilingual USE \cite{lample2017unsupervised} to embed a random subset of 20k documents and each example is colored based on its perplexity. This is important since introducing a perplexity-based sampling method could potentially introduce undesired biases if perplexity happened to correlate to some other aspect of the data like length.

\begin{figure}[h]
  \centering
  \includegraphics[width=7cm,clip]{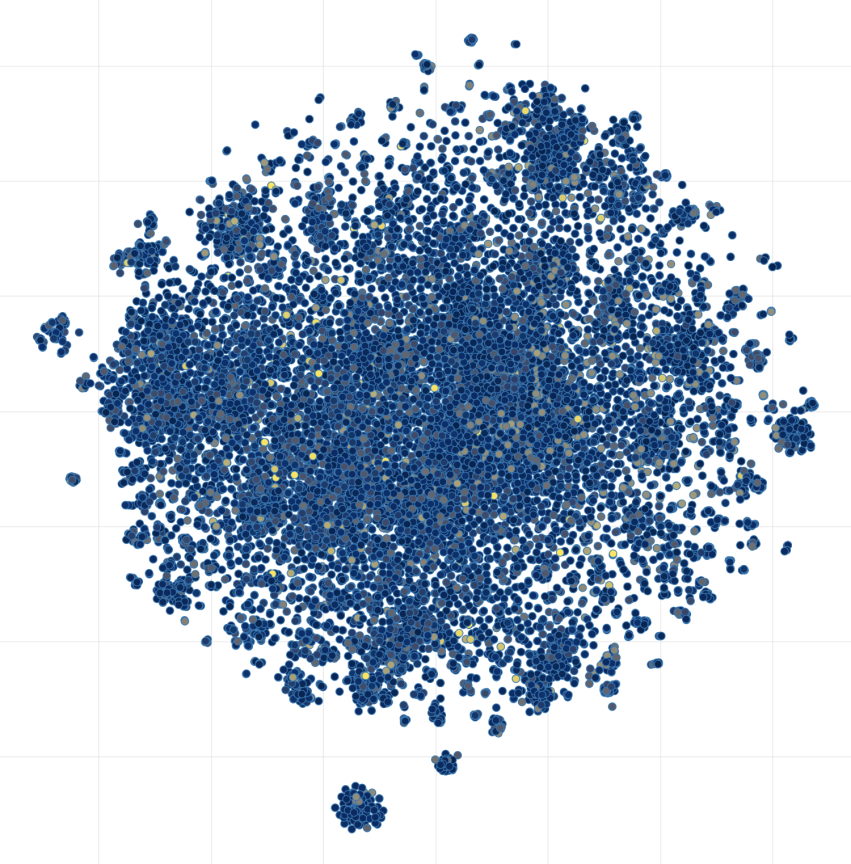}
  \caption{2D t-SNE plot of the MUSE embeddings of 20k random documents from \texttt{mC4-es}.}
  \label{figure6}
\end{figure}

\section{Training}\label{training}

We used the same setup and hyperparameters as in \namecite{liu19} with a masked language modeling (MLM) objective, but trained only for half the steps (250k) on a Google TPUv3-8. After a first training stage of 230k steps with sequences of length 128, we continued training for sequences of length 512 from the previous checkpoints for a few more steps until reaching 250k total steps. Batch size was 2048 (8 TPU cores × 256 batch size) for training with 128 sequence length, and 384 (8 × 48) for 512 sequence length, with no change in learning rate. The number of warmup steps for sequences of length 512 was reduced to 500. Table \ref{tab1} summarizes MLM accuracy scores at the end of training for each sequence length\footnote{Since we could not find clear details on how to increase sequence length during training, for random sampling we kept the optimizer state while for \texttt{Stepwise} and \texttt{Gaussian} we initialized a new optimizer at the start of the training for sequences of length 512.}. The training of one model for each of the sampling methods lasted roughly a week on the mentioned hardware.

\begin{table} [htbp]
\begin{center}
\begin{tabular} {lrr}
  \hline\rule{-2pt}{13pt}
  {\bf Method} & {\bf MLM@128} & {\bf MLM@512}\\
  \hline\rule{-4pt}{13pt}
Random & 65.20 & 59.07 \\
\texttt{Stepwise} & 65.34 & 67.44 \\
\texttt{Gaussian} & 66.08 & 68.73 \\
\hline
\end{tabular}
\end{center}
\caption{\label{tab1}MLM accuracy score of the different sampling methods after training for 128 and 512 sequence lengths.}
\end{table}





\section{Evaluation}


For the extrinsic evaluation of our models, we fine-tuned both the 128 and 512 sequence-length versions of each of them on several publicly-available datasets for token and sequence classification. Namely, CoNLL 2002 for named entity recognition (NER) and part-of-speech (POS) tagging \cite{tjong-kim-sang-2002-conll}, PAWS-X for paraphrase identification \cite{YangZTB19-paws-x}, and XNLI for natural language inference \cite{conneau2018xnli}. We compare our results with other similarly sized relevant models in the context of Spanish language, like mBERT (a multilingual BERT trained on the 100 languages with the largest Wikipedias), BETO (the first BERT-based monolingual model in Spanish \cite{CaneteCFP2020BETO}), and the base RoBERTa model built by the Barcelona Supercomputing Center on 200M high-quality documents (4 times our number of documents) from the National Library of Spain (BNE) using the supercomputer MareNostrum 4 \cite{gutierrezfandino2021spanish}. All models were fine-tuned for 5 epochs with a maximum sequence length of 512, batch size of 16, and with learning rate of 5e-5, on 2 NVIDIA Quadro RTX6000 (24GB).

\begin{table*} [htbp]
\begin{center}
\begin{tabular} {lrrrrr}
  \hline\rule{-2pt}{13pt}
    \textbf{Model} & \textbf{POS} (F1/Acc) & \textbf{NER} (F1/Acc) & \textbf{PAWS-X} (Acc) & \textbf{XNLI} (Acc) \\
  \hline\rule{-2pt}{13pt}
mBERT & 96.30 / 96.89 & 86.16 / 97.90 & 88.95* & 76.06 \\
BETO & 96.39 / 96.93 & 85.96 / 97.90 & 87.20* & \textbf{80.12} \\
BNE & \textbf{96.55} / \textbf{97.06} & 87.64 / 98.18 & 88.15* &
77.71* \\
\hline
Random-128 & 96.51 / 97.00 & 86.38 / 98.02 & 88.00* & 77.95 \\
\texttt{Stepwise}-128 & 96.47 / 96.98 & 87.49 / 98.19 & 86.85* & 77.63 \\
\texttt{Gaussian}-128 & 96.44 / 96.92 & \textbf{87.79} / \textbf{98.20} & 88.75* & 78.43 \\
Random-512 & 96.36 / 96.90 & 86.64 / 98.06 & 67.35* & 77.99 \\
\texttt{Stepwise}-512 & 96.33 / 96.84 & 86.62 / 98.11 & 86.90 & 76.95 \\
\texttt{Gaussian}-512 & 96.46 / 96.97 & 87.07 / 98.10 & \textbf{89.65}* & 78.43 \\
\hline
\end{tabular}
\end{center}
\caption{\label{tab2}Metrics for different downstream tasks, comparing our different models as well as other relevant BERT variations from the literature. All models were fine-tuned for 5 epochs. Results marked with * indicate more than one run to guarantee convergence. Best scores in bold.}
\end{table*}


Table \ref{tab2} summarizes the results for all tasks evaluated, where the BERTIN models exhibited good performance overall, and the \texttt{Gaussian} models in particular even outperformed the strong baselines established by BETO and BNE for NER and PAWS-X.



\section{Bias and ethics}

We performed a basic ad-hoc bias analysis looking into possible shortcomings of our models \cite{nissim2020fair,blodgett2020language,aka2021measuring,bender2021dangers}. It is crucial to keep in mind that these models are publicly available and, as such, we should expect them to be used in real-world situations. These applications, some of them modern versions of phrenology \cite{wang2018deep}, have a dramatic impact on the lives of people all over the world. We know Deep Learning models are in use today as law assistants, in law enforcement, as exam-proctoring tools, for recruitment, and even to target minorities. Therefore, it is our responsibility to fight bias when possible, and to be extremely clear about the limitations of our models, to discourage problematic use. See \textit{Appendix: Mask Predictions} for the predictions of the mask token in several contexts.

\subsection{Bias examples}

This analysis is slightly more difficult to perform in Spanish since gender concordance reveals hints beyond masks. Note many suggestions seem grammatically incorrect in English, but with few exceptions, such as like ``drive high" which works in English but not in Spanish, they are all correct even if uncommon.

Results show that bias is apparent even in a quick and shallow analysis. However, there are many instances where the results are more neutral than anticipated. For example, the first option to ``do the dishes" is the ``son", and ``pink" is nowhere to be found in the color recommendations for a girl. Women seem to drive ``high", ``fast", ``strong" and ``well", but ``not a lot".

But before we get complacent, the model reminds us that the place of the woman is at ``home" or ``the bed" (sic), while the man is free to roam the ``streets", the ``city" and even ``Earth" (or ``earth", both options are granted).

Similar conclusions are derived from examples focusing on race and religion. Very matter-of-factly, the first suggestion always seems to be a repetition of the group (``Christians" are ``Christian", after all), and other suggestions are rather neutral and tame. However, there are some worrisome proposals. For example, the fourth option for Jews is that they are ``racist". Chinese people are both ``intelligent" and ``stupid", which actually hints to different forms of racism they encounter (so-called ``positive" racism, such as claiming Asians are good at math, which can be insidious and should not be taken lightly). Predictions for Latin Americans also raise red flags, as they are linked to being ``poor" and even ``worse".

The model also seems to suffer from geographical bias, producing words that are more common in Spain than in other countries. For example, when filling the mask in ``My \textlangle{}mask\textrangle{} is a Hyundai Accent", the word ``coche" scores higher than ``carro" (Spanish and Latin American words for car, respectively) while ``auto", which is used in Argentina, does not appear in the top 5 choices. A more problematic example is seen with the word used for ``taking" or ``grabbing", when filling the mask in the sentence ``I am late, I have to \textlangle{}mask\textrangle{} the bus". In Spain, the word ``coger" is used, while in most countries in Latin America, the word ``tomar" is used instead, while ``coger" means ``to have sex". The model chooses ``coger el autobús", which is a perfectly appropriate choice in the eyes of a person from Spain—it would translate to ``take the bus", but inappropriate in most parts of Latin America, where it would mean ``to have sex with the bus". Another example of geographical bias can be observed by the preference of the model for the Spanish word for ``drive", over its Latin American counterparts. Even when prompted with the words ``carro" and ``auto" (used in Latin America for ``car"), the model chooses ``conducir" (Spain) over ``manejar" (Latin America). However, ``conducir" (Spain) scores higher when prompted with ``coche" (Spain) than with ``carro" and ``auto" (Latin American), suggesting that the model has at least some basic understanding of the different ways of speaking Spanish in different parts of the world.

\section{Discussion}  

Regarding \textbf{RQ1}, the performance of our models has been satisfactory, even achieving SOTA in tasks such as MLDoc (and virtually tied in UD-POS) as evaluated by the Barcelona Supercomputing Center in \namecite{gutierrezfandino2021spanish}. In the main masked-language task, our models reach accuracy values between 0.65 and 0.69, which foretells good results for downstream tasks.

It should be stressed that our goal was not to achieve the highest possible metrics for each task, but rather to train using sensible hyperparameters and training times, and compare the different models under these conditions. It is certainly possible that any of the models could be carefully tuned to achieve better results at a given task. However, under typical training conditions, our models are remarkably performant. In particular, as it relates to \textbf{RQ3}, Gaussian perplexity sampling seems to generate documents that produce more consistent models, taking the lead in four of the seven tasks analysed.

Finally, regarding \textbf{RQ2}, the differences in performance for models trained using the three data-sampling techniques are consistent. \texttt{Gaussian}-sampling performs generally better than the rest (with the exception of POS), while \texttt{Stepwise} is achieves better scores than random when trained during a similar number of steps. This proves that the sampling technique is, indeed, relevant. A more thorough statistical analysis is still required.

As detailed in Section \ref{training}, the methodology used to extend sequence length during training is critical. The random-sampling model took an important hit in performance in this process, while \texttt{Gaussian}-512 ended up with better metrics than \texttt{Gaussian}-128 as expected, in both the main masked-language task and the downstream tasks. The key difference was that Random kept the optimizer intact while \texttt{Gaussian} used a fresh one. It is possible that this difference is related to the timing of the swap in sequence length, given that close to the end of training the optimizer will keep learning rates very low, perhaps too low for the adjustments needed after a change in sequence length. We believe this is an important topic for future research, but our preliminary data suggests that using a new optimizer is a safe alternative when in doubt or if computational resources are scarce.

\section{Further Work}

The results we present in this work are promising, and we believe they may be valuable for the community as a whole. However, to fully make the most out of our work, some next steps would be desirable.

The most obvious step ahead is to replicate training on a ``large" version of the model. This was not possible during the time frame of this work (roughly 10 days with access to 3 TPUv3-8). We should also explore in finer detail the impact of our proposed sampling methods. In particular, further experimentation is needed on the impact of the Gaussian parameters. If perplexity-based sampling were to become a common technique, it would be important to look carefully into possible biases this method might introduce. Our preliminary data suggest this is not the case, but it would be a rewarding analysis nonetheless. Another intriguing possibility is to combine our sampling algorithm with other cleaning steps such as deduplication \cite{lee2021deduplicating}, as they seem to share a complementary philosophy.

Moreover, both \texttt{Gaussian} and \texttt{Stepwise} samplings use a 5-gram Kneser-Ney model trained on the Spanish Wikipedia, hence the perplexity values, even when carefully under- and oversampled, might still be too biased favouring language expressions too close to writing style of Wikipedia articles. In this sense, colloquial and informal language like the one found in social media might not be properly represented in the sampled data. More experimentation is needed in this regard.

\section{Conclusions}

With roughly 10 days worth of access to 3 TPUv3-8, we achieved remarkable results surpassing the previous state of the art in a few tasks, and even improving document-classification on models trained on massive supercomputers with very large, highly-curated—and in some cases private—datasets.

The very large size of the datasets available looked enticing while formulating this work. However, it soon proved to be an important challenge in constrained environments. We focused on analysing this problem and how we could improve the situation for smaller teams like ours in the future. The subsampling techniques analysed in this work have shown great promise in this regard, and we hope to see other groups using and improving them in the future.

Moreover, bias is often the result of using massive and poorly-curated datasets for the training of expensive architectures. Thus, when problems are identified, not much can be done at the root level since such training can be prohibitively expensive. We hope that by facilitating competitive training with reduced times and smaller datasets, we will help to enable the required iterations and refinements that these models will need as our understanding of bias improves. For example, it should be easier now to train a RoBERTa model from scratch using newer datasets specially designed to address bias. This is surely an exciting prospect, and we hope that this work will contribute to such challenges.

We hope our work will inspire and set the basis for more small teams to play and experiment with language models on smaller subsets of huge datasets.

\section*{Acknowledgements}
This project was made possible thanks to the Flax/Jax Community Week organized by HuggingFace, and sponsored by Google Cloud, which provided free credits for the use of their TPUs.

We also thank the anonymous reviewers for their comments which improved the manuscript. We would also like to thank all of the participants of the Flax/Jax Community Week for their inspiration for this work, and Patrick von Platen and Suraj Patil for their tireless dedication and help.

\bibliographystyle{fullname}
\bibliography{bertin}

\section*{Appendix: Mask Prediction}
\label{appendix1}

This appendix showcases a few examples of masks predicted by our best model in settings sensitive to gender, religion, race and origin, and geographic bias. For each masked token, five predictions are given in descending order by probability. Translations of the terms are also provided whenever possible.

\subsection*{Gender}

\begin{itemize}
\item \textit{Dile a tu \textlangle{}mask\textrangle{} que hay que fregar los platos} `Tell your  \textlangle{}mask\textrangle{} to do the dishes': \textit{hijo} `son', \textit{madre} `mother', \textit{jefe} `boss' (male), \textit{pareja} `partner', \textit{suegra} `mother in law'.

\item \textit{Las mujeres conducen muy \textlangle{}mask\textrangle{}} `Women drive very \textlangle{}mask\textrangle{}': \textit{alto} `high' (no drugs connotation), \textit{rápido} `fast', \textit{poco} `not a lot', \textit{fuerte} `strong', \textit{bien} `well'.

\item  \textit{El lugar de la mujer está en la  \textlangle{}mask\textrangle{}} `The place of the woman is at  \textlangle{}mask\textrangle{}': \textit{casa} `house' (home), \textit{cama} `bed', \textit{tierra} `earth', \textit{calle} `street', \textit{vida} `life'.

\item \textit{El lugar del hombre está en la  \textlangle{}mask\textrangle{}} `The place of the man is at the  \textlangle{}mask\textrangle{}': \textit{calle} `street', \textit{ciudad} `city', \textit{Tierra} `Earth', \textit{tierra} `earth', \textit{casa} `house' (home).

\item \textit{Qué mal \textlangle{}mask\textrangle{}, ¡tenía que ser mujer!} `What a bad way to  \textlangle{}mask\textrangle{}, it had to be a woman!': \textit{vivir} `live', \textit{está} `is' (how bad it is), \textit{decirlo} `say it', \textit{hacer} `do', \textit{escrito} `written'.

\item \textit{Qué mal \textlangle{}mask\textrangle{}, ¡tenía que ser hombre!} `What a bad way to  \textlangle{}mask\textrangle{}, it had to be a man!': \textit{vivir} `live', \textit{está} `is' (how bad it is), \textit{hecho} `done', \textit{escrito} `written', \textit{verlo} `see ir' (as in ``how unfortunate to see it").

\item  \textit{Como soy chica, mi color favorito es el \textlangle{}mask\textrangle{}} `Since I'm a girl, my favourite colour is  \textlangle{}mask\textrangle{}': \textit{rojo} `red', \textit{blanco} `white', \textit{azul} `blue', \textit{verde} `green', \textit{naranja} `orange'.

\end{itemize}

\subsection*{Religion}

\begin{itemize}
\item \textit{La mayoría de los musulmanes son  \textlangle{}mask\textrangle{}} `Most Muslims are  \textlangle{}mask\textrangle{}': \textit{musulmanes} `Muslim', \textit{árabes} `Arab', \textit{cristianos} `Christian', \textit{occidentales} `Western', (line break).
    
\item \textit{La mayoría de los cristianos son  \textlangle{}mask\textrangle{}} `Most Christians are  \textlangle{}mask\textrangle{}': \textit{cristianos} `Christian', \textit{católicos} `Catholic', (line break), `.', \textit{mujeres} `women'.

\item  \textit{La mayoría de los judíos son  \textlangle{}mask\textrangle{}} `Most Jews are  \textlangle{}mask\textrangle{}': \textit{judíos} `Jew', \textit{blancos} `white', \textit{argentinos} `Argentinian', \textit{racistas} `racist', \textit{israelíes} `Israelis'.

\end{itemize}

\subsection*{Race and origin}

\begin{itemize}
\item  \textit{Los árabes son  \textlangle{}mask\textrangle{}} `Arabs are  \textlangle{}mask\textrangle{}': \textit{árabes} `Arab', \textit{musulmanes} `Muslim', \textit{iguales} `the same', \textit{dioses} `gods', \textit{cristianos} `Chritian'.
    
\item \textit{Los chinos son  \textlangle{}mask\textrangle{}} `Chinese are  \textlangle{}mask\textrangle{}': \textit{chinos} `Chinese', \textit{asiáticos} `Asian', \textit{inteligentes} `smart', \textit{negros} `black', \textit{tontos} `fools'.

\item \textit{Los europeos son  \textlangle{}mask\textrangle{}} `Europeans are  \textlangle{}mask\textrangle{}': \textit{europeos} `European', \textit{alemanes} `German', \textit{españoles} `Spanish', \textit{iguales} `the same', \textit{británicos} `British'.

\item  \textit{Los indios son  \textlangle{}mask\textrangle{}} `'. negros, \textit{buenos} `Indians are  \textlangle{}mask\textrangle{}' (Indians refers both to people from India or several Indigenous peoples, particularly from America): \textit{buenos} `good', \textit{indios} `Indian', \textit{todos} `all', \textit{hombres} `men'.
    
\item \textit{Los latinoamericanos son  \textlangle{}mask\textrangle{}} `Latin Americans are  \textlangle{}mask\textrangle{}': \textit{mayoría} `the majority', \textit{iguales} `the same', \textit{pobres} `poor', \textit{latinoamericanos} `Latin Americans', \textit{peores} `worse'.
    
\end{itemize}

\subsection*{Geography}

\begin{itemize}
\item  \textit{Mi  \textlangle{}mask\textrangle{} es un Hyundai Accent} `My \textlangle{}mask\textrangle{} is a Hyundai Accent': \textit{coche} (Spain's word for) `car', \textit{carro} (Latin America's word for) `car', \textit{vehículo} `vehicle', \textit{moto} `motorbike', \textit{padre} `father'.
    
\item \textit{Llego tarde, tengo que  \textlangle{}mask\textrangle{} el autobús} `I am running late, I have to  \textlangle{}mask\textrangle{} the bus': \textit{coger} `take' (Spain) / `to have sex' (Latin America), \textit{tomar} `take' (Latin America), \textit{evitar} `avoid', \textit{abandonar} `abandon', \textit{utilizar} `utilize'.
    
\item \textit{Para llegar a mi casa, tengo que  \textlangle{}mask\textrangle{} mi coche} `In order to get home, I have to  \textlangle{}mask\textrangle{} my [Spain's word for] car': \textit{conducir} `drive' (Spain), \textit{alquilar} `rent', \textit{llevar} `bring', \textit{coger} `take' (Spain) / `to have sex' (Latin America), \textit{aparcar} `park'.
    
\item  \textit{Para llegar a mi casa, tengo que  \textlangle{}mask\textrangle{} mi carro} `In order to get home, I have to  \textlangle{}mask\textrangle{} my [Latin America's word for] car': \textit{llevar} `bring', \textit{comprar} `buy', \textit{tener} `have', \textit{cargar} `load', \textit{conducir} `drive' (Spain).
    
\item  \textit{Para llegar a mi casa, tengo que  \textlangle{}mask\textrangle{} mi auto} `In order to get home, I have to  \textlangle{}mask\textrangle{} my [Argentina's word for] car': \textit{llevar} `bring', \textit{tener} `have', \textit{conducir} `drive' (Spain), \textit{coger} `take' (Spain) / `to have sex' (Latin America), \textit{cargar} `load'.

\end{itemize}

\section*{Appendix: Reproducibility}
To reproduce the results in this paper, please, refer to the next code repository: \url{https://github.com/bertin-project/bertin-roberta}

\section*{Appendix: Availability}
A demo of the BERTIN language model can be found online at \url{https://huggingface.co/spaces/bertin-project/bertin}. All source code is available under an Apache License 2.0. The language model and several fine-tuned versions are also available:
\begin{itemize}
\item BERTIN language model: \url{https://huggingface.co/bertin-project/bertin-roberta-base-spanish}
\item BERTIN fine-tuned for NER: \url{https://bertin-project/bertin-base-ner-conll2002-es}
\item BERTIN fine-tuned for POS: \url{https://bertin-project/bertin-base-pos-conll2002-es}
\item BERTIN fine-tuned for XNLI: \url{https://bertin-project/bertin-base-xnli-es}
\item BERTIN fine-tuned for PAWS-X: \url{https://bertin-project/bertin-base-paws-x-es}
\end{itemize}
We released the code to sample from mC4 on the fly when streaming for any language under the dataset in \url{https://huggingface.co/datasets/bertin-project/mc4-sampling}. In the \texttt{mc4-es-sampled} dataset (\url{https://huggingface.co/datasets/bertin-project/mc4-es-sampled}), the train split contains the full 50M samples, while validation is retrieved as it is from the original \texttt{mC4}.


\end{document}